\documentclass{bmvc2k}
\usepackage{amsmath}
\usepackage{amssymb}
\usepackage{multicol}
\usepackage{graphicx}
\usepackage{epstopdf}
\usepackage{pbox}
\usepackage[normalem]{ulem} % either use this (simple)
\DeclareMathOperator*{\argmin}{argmin}

\title{Sparse vs. Non-sparse: Which One Is Better for Practical Visual Tracking?}

\addauthor{Yashar Deldjoo}{yashar.deldjoo@polimi.it}{1}
\addauthor{Shengping Zhang}{shengping.zhang@gmail.com}{2}
\addauthor{Bahman Zanj}{zanj@guilan.ac.ir}{3}
\addauthor{Paolo Cremonesi}{paolo.cremonesi@polimi.it}{1}
\addauthor{Matteo Matteucci}{matteo.matteucci@polimi.it}{1}

\addinstitution{
 Via Ponzio 34/5\\
 Politecnico di Milano\\
 Milano, Italy
}

\addinstitution{
 224 Waterloo Road,\\
 Harbin Institute of Technology,\\
 Hong Kong, China
}
\addinstitution{
 Faculty of Engineering\\
 University of Guilan,\\
 Rasht, Iran
}

\runninghead{}{}

\begin{document}

\maketitle

\begin{abstract}
Recently, sparse representation  based visual tracking methods have attracted increasing attention in the computer vision community. Although achieve superior performance to traditional tracking methods, however, a basic problem has not been answered yet --- that whether the sparsity constrain is really needed for visual tracking? To answer this question, in this paper, we first propose a robust non-sparse representation based tracker and then conduct extensive experiments to compare it against several state-of-the-art sparse representation based trackers. Our experiment results and analysis indicate that the proposed non-sparse tracker achieved competitive tracking accuracy with sparse trackers while having faster running speed, which support our non-sparse tracker to be used in practical applications. 
%Recently, models based on Sparse Representation (SR) that represent a target as a superposition of a suitable updating dictionary have attracted attentions in visual research community for Visual Object Tracking (VOT). 
%Different visual tracking methods based on SR and its variants such as structured sparsity have been proposed in recent years.
%While there is a desire to propose the effectiveness of SR for accurate VOT, the underlying driving factor for precise and robust VOT has not been well investigated. 
%The recent trend on sparse-based VOT borrows main ideas from Face Recognition (FR) research, using over--complete dictionaries containing both target and occlusion templates. 
%We argue that the two recognition problem of FR and VOT are fundamentally different problems. 
%We illustrate the differences between sparse-based FR and VOT and show that sparse-based VOT essentially is a low-dimensional linear subspace modeling. 
%Thus, statistical Least Square Estimators (LSEs) not only can provide a pretty competitive precise results but also has much more computational efficiency. 
%To this end, a comprehensive analysis is presented with extensive experiments on several challenging video tracking sequences to show that LSEs can be used for accurate real-time visual tracking.
\end{abstract}

%------------------------------------------------------------------------- 
\section{Introduction}
\label{sec:intro}
Visual tracking, \emph{i.e.}, tracking a specific target object in consecutive video frames to get its moving trajectory, is one of the most important tasks in computer vision. A wide range of applications rely on robust visual tracking including, security and surveillance~\cite{Hu:2004:SVS:2220414.2220805,Kim2010}, vehicle transportation and traffic monitoring~\cite{Coifman1998,Kastrinaki2003,Atev:2005:VAC:2218579.2218774}, video compression~\cite{Mitchell:1996:MVC:548218, Sikora:1997:MVS:2322483.2322735, Hariharakrishnan:2005:FOT:2219086.2219477}, head-tracking, gesture recognition and eye-gaze tracking~\cite{deldjoo2016low,deldjoo2009wii,Pavlovic:1997:VIH:261506.272696, Al-rahayfeh2013}. Visual tracking has been extensively studied in the past decades in the computer vision community; however, it is still very challenging to handle irregular appearance changes of the tracked object during tracking, which are mainly due to abrupt geometric transformation, photometric variations like sudden change in illumination, and partial or full occlusions.
 
In the literature, a large number of tracking approaches have been proposed which can be roughly grouped in two main classes: \textit{discriminative} methods and \textit{generative} methods. The former formulates the tracking problem as the binary classification of distinguishing the object from its background while the latter builds an appearance model of the target and formulates the tracking problem as a matching problem. Recently, inspired by the success of sparse representation in face recognition~\cite{Wright:2009:RFR:1495801.1496037}, sparse coding~\cite{olshausen1996emergence} has been successfully used in visual tracking~\cite{Mei09,Zhang:2013:SCB:2445640.2446001,zhong2012robust,zhang2012robust,ji2015object}. Among them, $\ell_1$ minimization based tracking method~\cite{Mei09} formulates visual tracking as a reconstruction problem in a linear space where it is reasonable to impose a \emph{sparse} constrain on the representation coefficients that  the tracked target should be linearly represented by a small set of target templates with small reconstruction error. To make the tracker robust to occlusions, a set of occlusion templates are used in the linear representation to handle occlusions. Since this pioneer work, several researchers have tried to improve it by constraining the activation of these extra templates to improve the tracking accuracy~\cite{bao2012real} or by reducing the dimension of space to reduce tracking computational complexity~\cite{Li11CVPR,zhang2012real}. In the following part of the paper, we name all $\ell_1$  minimization based trackers as $\ell_1$ trackers.

Although promising results were reported at the time~\cite{Mei09} was written and even though a number of other works have applied $\ell_1$ trackers in their specific contexts~\cite{Zhang:2013:SCB:2445640.2446001,zhong2012robust,zhang2012robust,ji2015object}, the real role of the sparse constrain in the sparse representation was not well investigated in videos containing a variety of tracking circumstances. In particular, several studies in object recognition~\cite{RigamontiCVPR2011,Zhang11ICCV} have experimentally indicated non-sparse representation with $\ell_2$ norm minization has gotten superior performance than sparse representation with $\ell_1$ norm mimization. Therefore, it is also necessary to investigate the roles of sparsity in $\ell_1$ trackers.  In addition, $\ell_1$ trackers are inevitably computationally expensive due to their iterative optimization procedure. Most $\ell_1$ trackers neglect the real-time requirement, which is very important for practical applications. In this paper, we aim at answering a basic question in $\ell_1$ trackers that whether sparsity is really needed for visual tracking. To this aim, we first propose a non-sparse tracker and then conduct extensive experiments to compare it against several sparse trackers. Our experiment results and analysis indicate that the proposed non-sparse tracker has achieved competitive tracking accuracy while having faster running speed, which is better than sparse trackers for practical applications. 
%argue that real-time performance and computational complexity lie at the core of tracking algorithms and they deserve a similar attention as to the other factors. In this line, this paper evaluate the roles of the sparsity constraint in $\ell_1$ trackers by conducting extensive experiments on 33 challenging  video sequences and evaluating different ways of constructing the dictionary  and imposing prior constraints on coefficients.  The result of such extended evaluation indicate xxxxx.

The rest of the paper is organized as it follows. Section~\ref{sec:sparse} first review the existing sparse trackers. In Section~\ref{sec:nonsparse}, the proposed non-sparse tracker is introduced in detail. Experiments are reported and analyzed in Section~\ref{sec:exp}. Section~\ref{sec:con} concludes the paper. 

%In section~\ref{sec:LSM}, we first introduce the general linear subspace modeling framework, based on which we review the $\ell_1$ trackers in section~\ref{sec:sparse} and present the proposed non-sparse tracker in section~\ref{sec:nonsparse}, respectively. In section~\ref{sec:exp}, extensive experiments were conducted to evaluate the sparse and non-sparse trackers. The conclusion was concluded in section~\ref{sec:con}

\section{Sparse tracker} \label{sec:sparse}
Inspired by the success of sparse representation in face recognition~\cite{Wright:2009:RFR:1495801.1496037}, Mei \emph{et al.} first proposed to model visual tracking as a sparse reconstruction problem under particle filter framework~\cite{Mei09} . In particular, let $\mathbf{y} \in \mathbb{R}^{d}$ be a feature vector obtained by stacking the pixel intensities of a target candidate into a column vector and $\mathbf{T} = \left[\mathbf{t}_1\ \mathbf{t}_2\ \ldots \mathbf{t}_{n} \right] \in \mathbb{R}^{d \times n}$  be the set of feature vectors of previous target templates, which is manually collected at the first frame and then updated in an online fashion over time.  It is natural to assume that the target templates $\mathbf{T}$ should span a linear space where the candidate is in. Formally, the target candidate $\mathbf{y}$ is represented in the following linear combination 
\begin{equation}
\label{eq:LM}
\mathbf{y} =  \alpha_1 \mathbf{t}_1 + \alpha_2 \mathbf{t}_2 + \ldots + \alpha_{n} \mathbf{t}_{n} + \boldsymbol{\eta} = \mathbf{T}\boldsymbol{\alpha} + \boldsymbol{\eta} 
\end{equation}
where the templates $\mathbf{T}$ constructs the sparse representation dictionary, $\boldsymbol{\alpha} = [\alpha_1, \alpha_2, \ldots, \alpha_n]^{\top}\in\mathbb{R}^n$ is the coefficient vector and $\boldsymbol{\eta}\in \mathbb{R}^d$ is the noise term. To handle occlusion, a set of occlusion templates $\mathbf{I} = [\mathbf{i}_1, \mathbf{i}_2, \ldots, \mathbf{i}_d]\in\mathbb{R}^{d\times d}$ is further introduced into the dictionary and the final linear combination is defined as
\begin{equation}
\label{eq:LM_augmented}
\mathbf{y} =  \begin{bmatrix} \mathbf{T} & \mathbf{I} \end{bmatrix}   \begin{bmatrix} \boldsymbol{\alpha} \\ \mathbf{e} \end{bmatrix}   = \mathbf{D}\mathbf{c}
\end{equation}
where a occlusion template $\mathbf{i}_i\in\mathbb{R}^d$ is a vector with only one nonzero entry (\emph{i.e.} $\mathbf{I}$ is an identity matrix), $\mathbf{D} = \begin{bmatrix} \mathbf{T} & \mathbf{I} \end{bmatrix} \in \mathbb{R}^{d\times (n+d)}$ is the augmented overcomplete dictionary, $\mathbf{c} = \begin{bmatrix} \boldsymbol{\alpha} \\ \mathbf{e} \end{bmatrix} \in \mathbb{R}^{n+d}$ is the augmented  coefficient vector and $\mathbf{e} = [e_1, e_2, \ldots, e_d]^{\top}\in\mathbb{R}^d$ is the occlusion coefficient vector. 

When assuming it to be sparse, the coefficient vector $\mathbf{c}$ can be obtained by solving the following $\ell_1$ minimization problem
\begin{equation}
\label{eq:l1norm}
\hat{\mathbf{c}} = \arg\min_{\mathbf{c}} \|\mathbf{y} - \mathbf{D}\mathbf{c}\|_2^2 + \lambda\|\mathbf{c}\|_1
\end{equation}
where the first and second terms measure the reconstruction error and the sparsity of the coefficient vector, respectively, and $\lambda$ is a constant that controls the importance of the reconstruction error to the sparsity. Once the coefficient vector is obtained, the tracking result is found as the target candidate with the smallest reconstruction error after projecting on the target template subspace, \emph{i.e.}, $\|\mathbf{y} - \mathbf{T}\boldsymbol{\alpha}\|_2^2$.  

\begin{figure}
\begin{center}
\begin{tabular}{ccc}
\bmvaHangBox{\includegraphics[width=3.45cm]{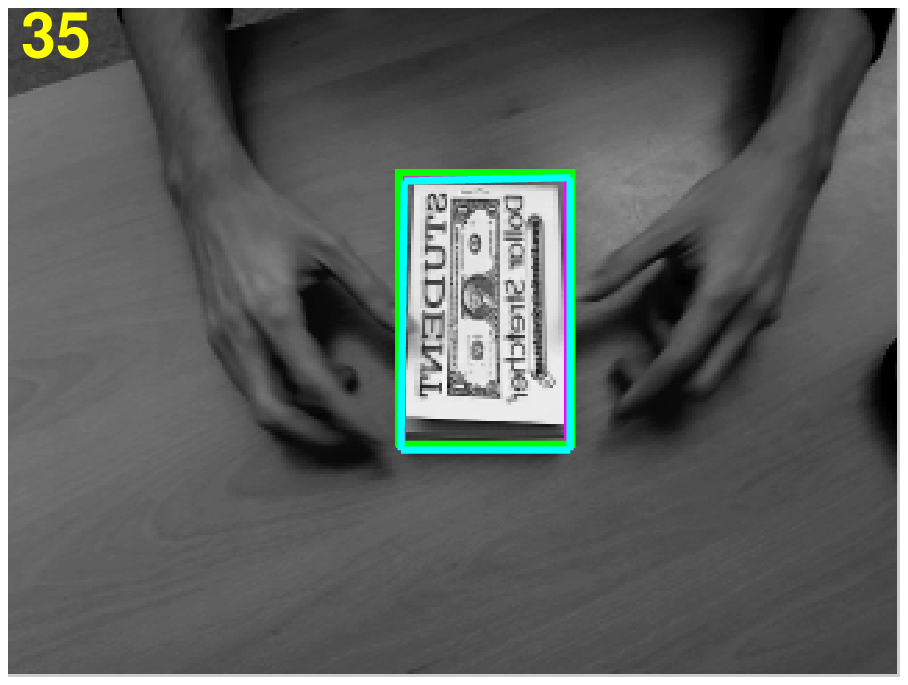}}& 
\bmvaHangBox{\includegraphics[width=3.45cm]{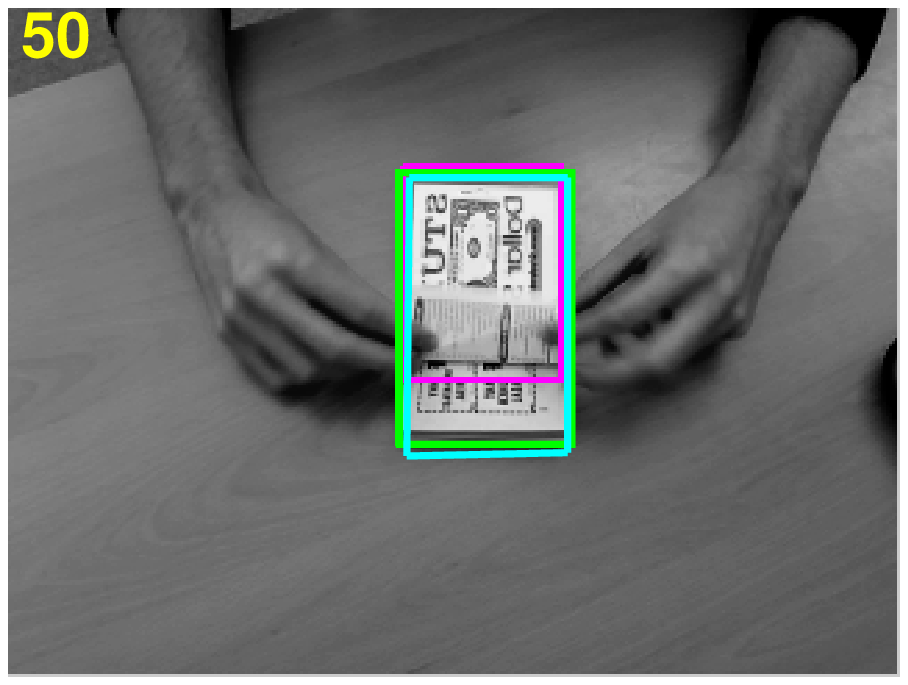}}&
\bmvaHangBox{\includegraphics[width=3.45cm]{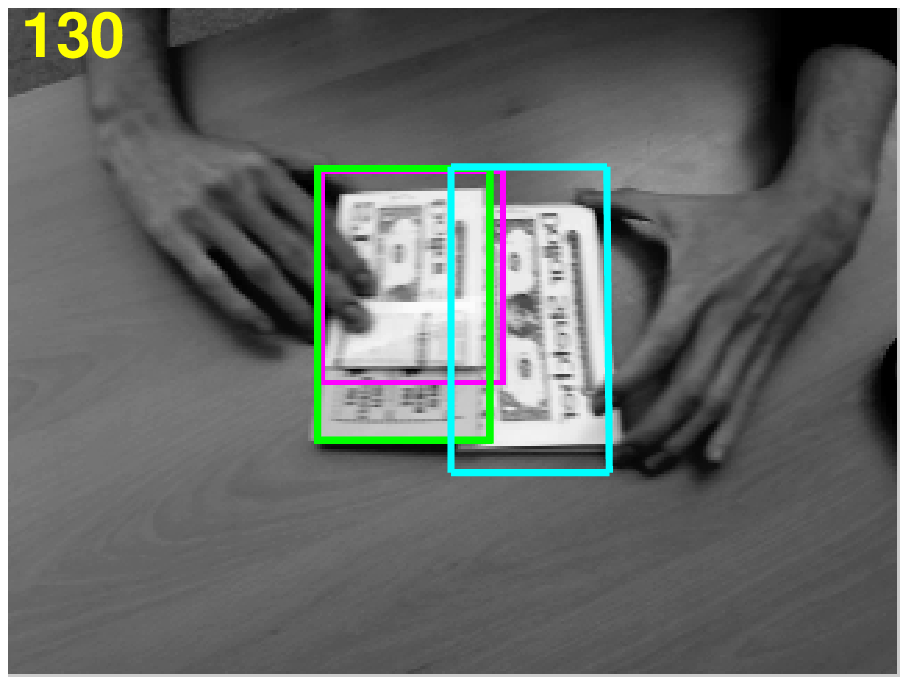}}\\ 
\vspace{1 mm}
\bmvaHangBox{\fbox{\includegraphics[width=3.25cm]{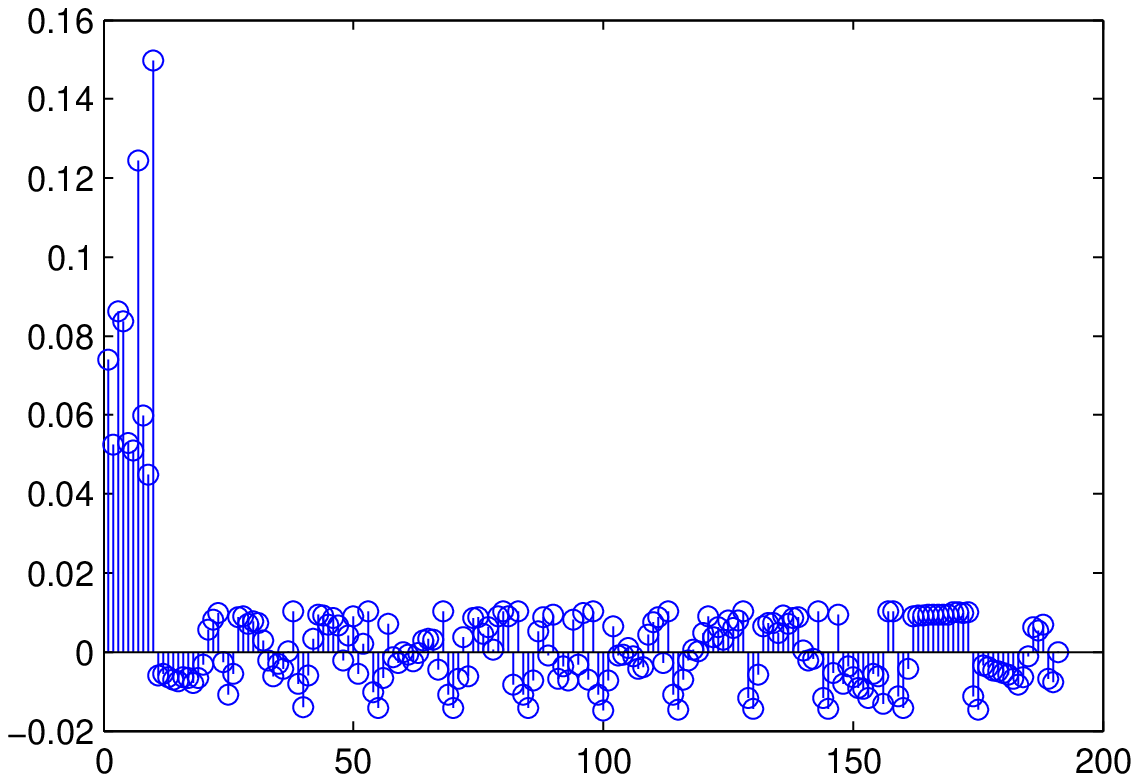}}}& 
\bmvaHangBox{\fbox{\includegraphics[width=3.25cm]{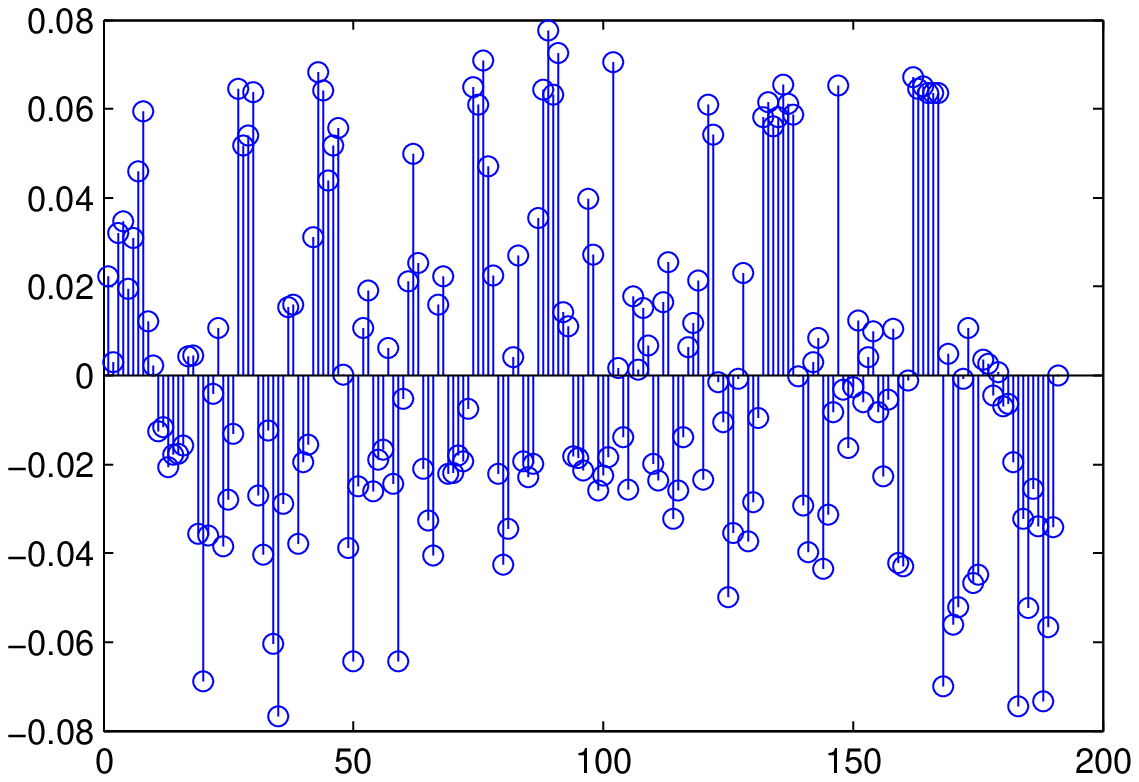}}}&
\bmvaHangBox{\fbox{\includegraphics[width=3.25cm]{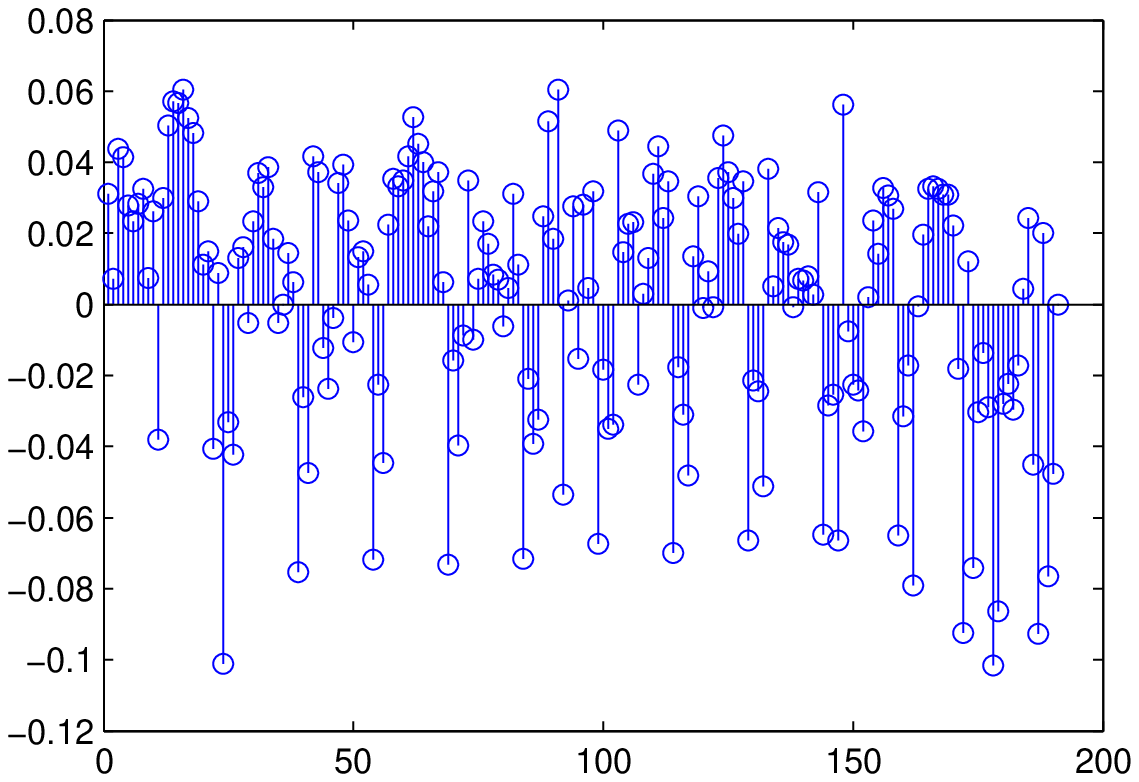}}}\\
\vspace{1.1 mm}
\bmvaHangBox{\fbox{\includegraphics[width=3.25cm]{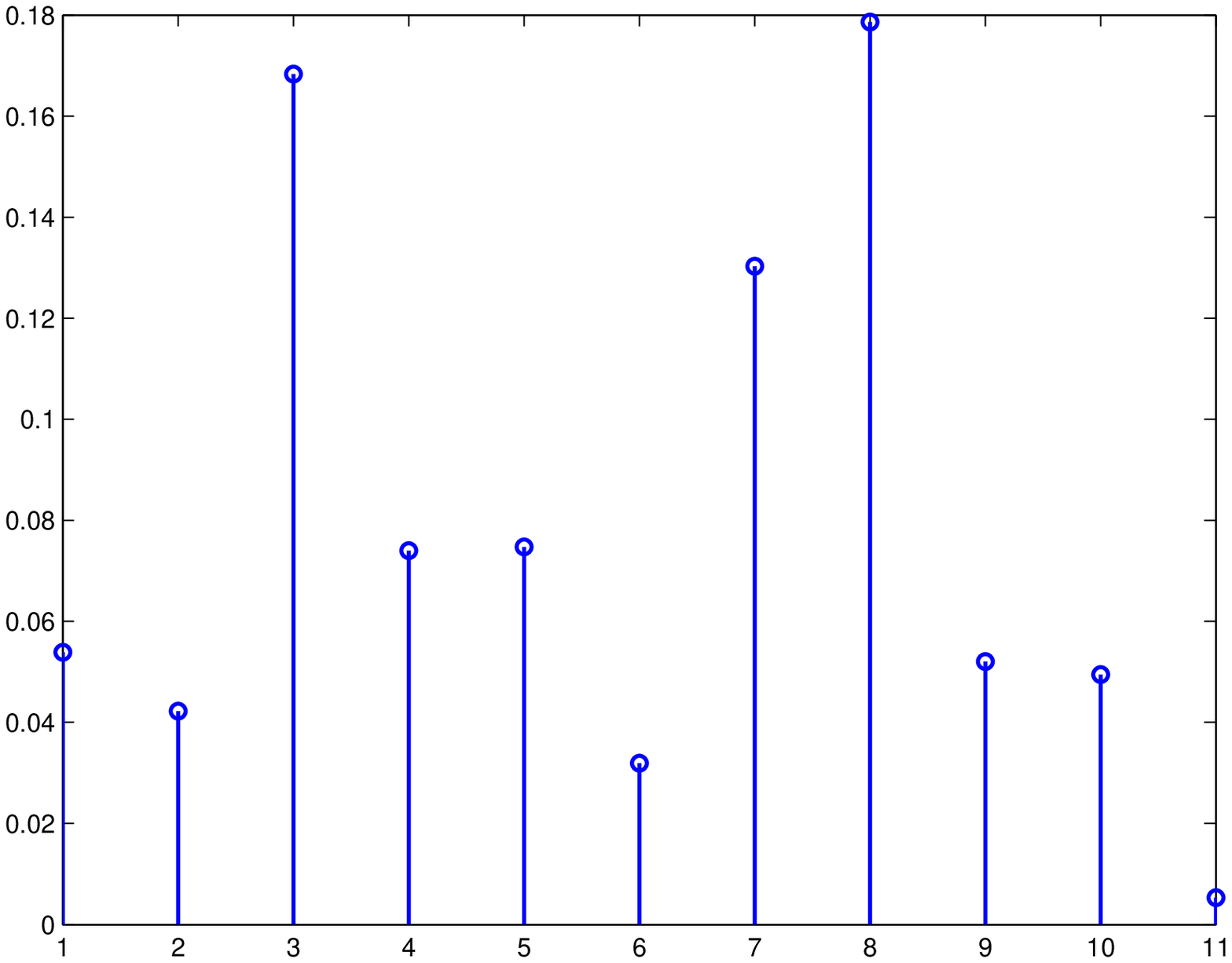}}}& 
\bmvaHangBox{\fbox{\includegraphics[width=3.25cm]{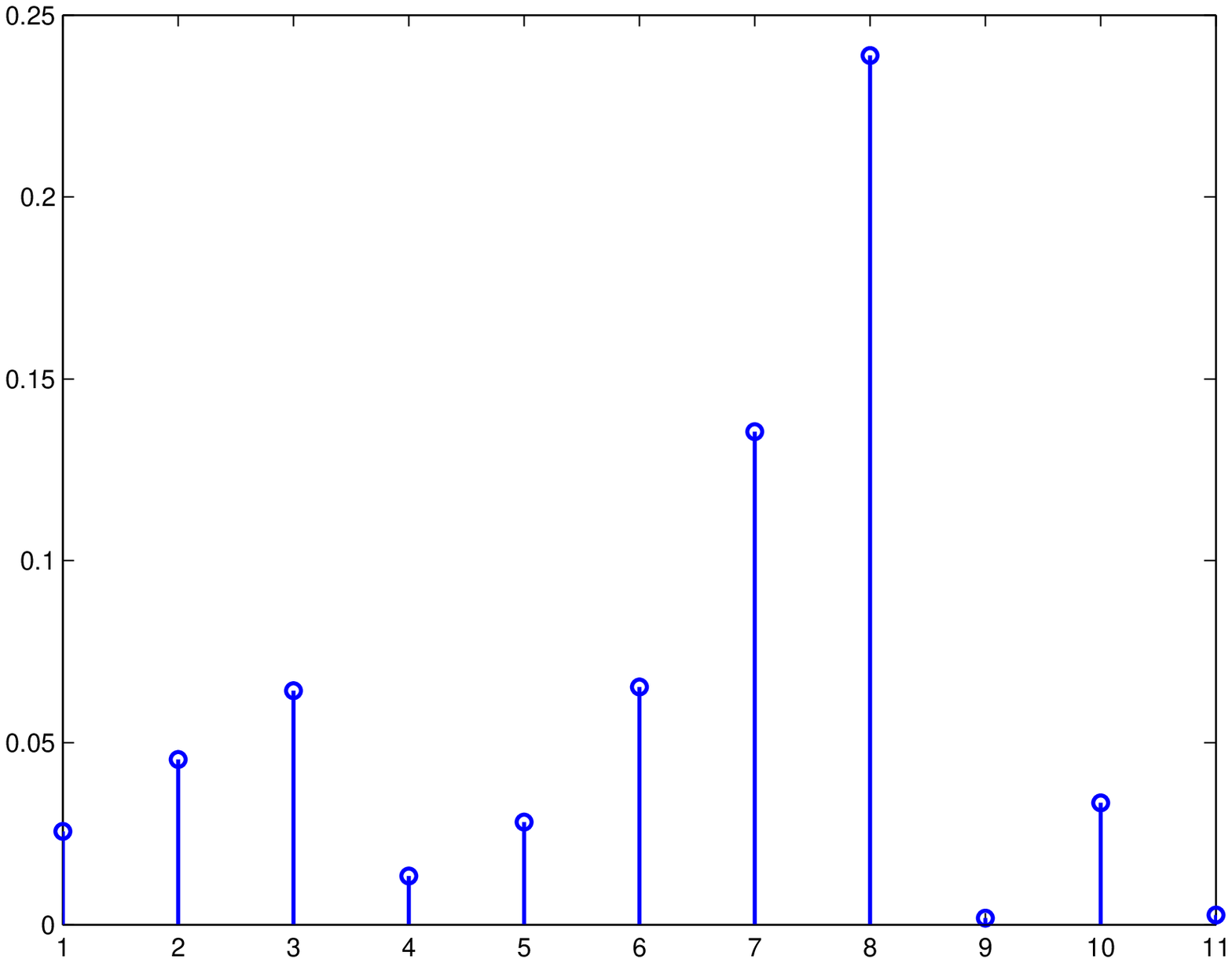}}}&
\bmvaHangBox{\fbox{\includegraphics[width=3.25cm]{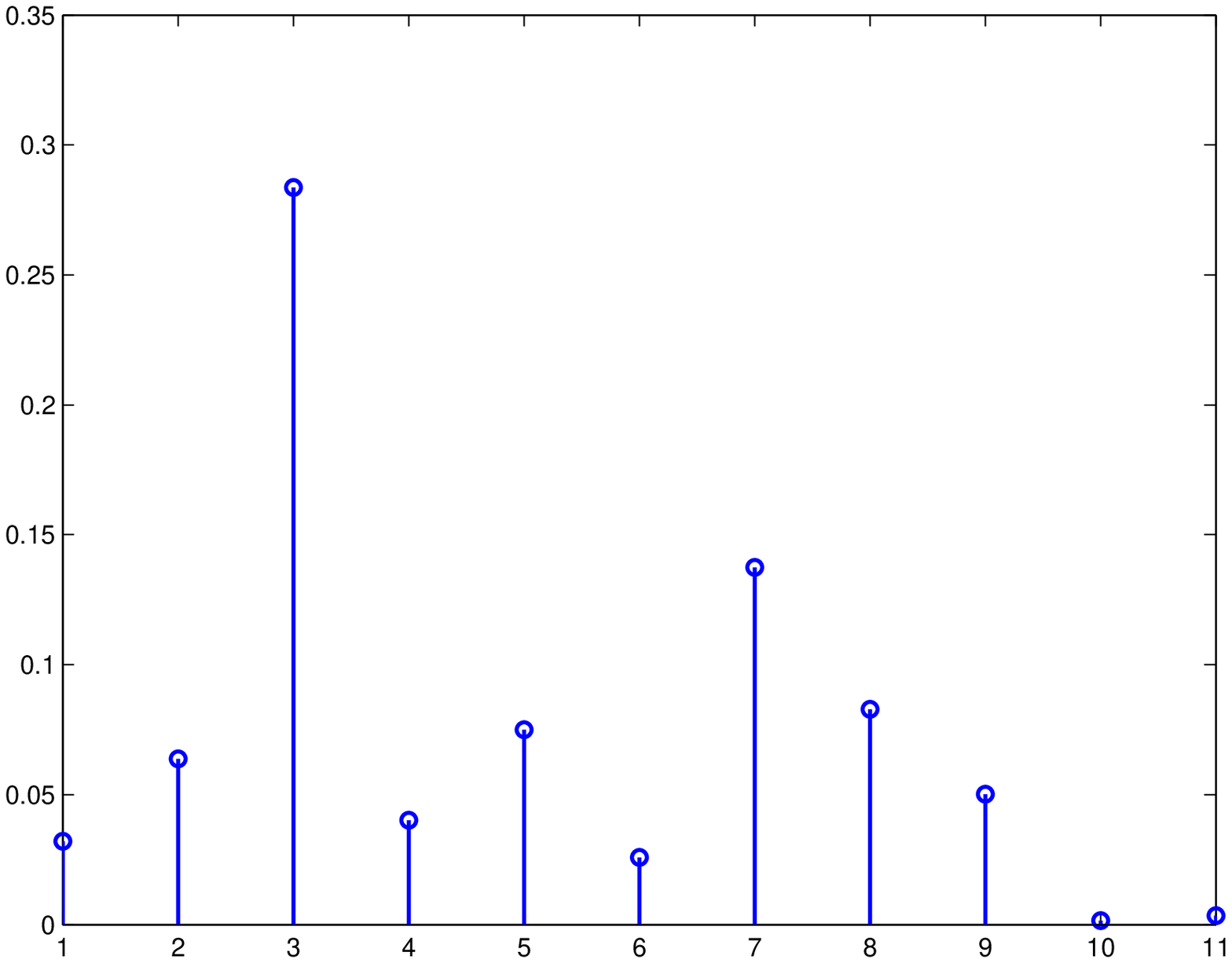}}}\\
\end{tabular}
\end{center}
\caption{Comparing visual tracking algorithms using different dictionary types, row 1: sample tracking frames - (green) baseline, (cyan) tracker using augmented template dictionary, (magenta) tracker using basis template dictionary, row 2: The solution coefficients for tracker cyan, row 3: The solution coefficients for tracker magenta.}
	%As can be seen the existence of occlusion templates leads to uncontrollable activation of them and drift of the tracker eventually. }
\label{fig1}
\end{figure}

	%\caption{\TODO{This caption is not clear. In the first row, there are only red, green and blue bounding boxes. Where is the black one?}Comparison of tracking methods using different prior constraints and dictionary type: (row 1) green:baseline, black: $\ell_{2}$ prior + basis dictionary, others: $\ell_{1}$ prior + augmented dictionary (row 2) The solution coefficients for $\ell_{1}$-tracker (row 3) The solution coefficients for $\ell_{2}$-tracker. 
	%As can be seen the existence of occlusion templates leads to uncontrollable activation of them and drift of the tracker eventually.
	%\label{fig1}
%\end{figure}
 
Although the desired performance was reported, especially the robustness to occlusion, there are several major drawbacks. Firstly, the sparse assumption on the coefficients may not hold in practice. In image classification field, several research studies~\cite{RigamontiCVPR2011,Zhang11ICCV} have indicated that non-sparse representation such as collaborative representation achieves the competitive classification performance with sparse representation. It is also necessary to investigate whether sparsity is really needed for visual tracking. Second, solving the $\ell_{1}$ minimization problem (Eq.~\ref{eq:l1norm}) is very time-consuming, which restricts the tracker being used in real-time.  Thirdly, the choice of occlusion templates is built upon a holistic idea to handle occlusion. In Figure~\ref{fig1}, we show an example how the use of an augmented dictionary containing occlusion templates can lead to target loss. The dollar notes have a similar appearance to the target note on top and as the person starts folding the note (frame 50) and moving it to the left (frame 130), in the model using augmented dictionary $\mathbf{D} = \left[\mathbf{T}, \mathbf{I}\right]$ we can see a large occlusion templates activated (\emph{i.e.} the coefficients become non-sparse) leading to the target loss whereas in simpler model with only basis target templates, the tracker learns the variation of the appearance in the target without trying to represent the difference via the help of occlusion templates as in the first case. We can conclude that the notion of occlusion templates to represent occlusion is built upon a holistic idea and in some cases it can lead to mis-classification of the target with its surrounding objects or background. To overcome the above drawbacks, several works have improved the work of~\cite{Mei09}.  For example, Mei \emph{et al.}~\cite{MeiMinimumCVPR2011} proposed to reduce the number of $\ell_1$-minimization by first sorting out the candidates based on their least-square residual error and accepting only candidates above a minimal threshold error to building a linear appearance model. Li \emph{et al.}~\cite{Li11CVPR} and Zhang et al.~\cite{zhang2012real} both made use of compressive sensing to build tracking models with real-time performance. An interesting work was proposed by Bao \emph{et al.}~\cite{bao2012real} in which the authors proposed a real-time $\ell_{1}$-tracker with improved tracking accuracy. The algorithm, which we shall revisit and refer to it as L1-APG, gains accuracy improvement via building a new minimization model for finding sparse representation of the target and real-time performance by a new APG (Accelerated Proximal Gradient) based numerical solver for resulting $\ell_{1}$ problem. 
\section{Non-Sparse ridge regression based tracker}
\label{sec:nonsparse}
In this paper, we propose a robust non-sparse tracker based on ridge regression (RR). Instead of augumenting occlusion templates in the dictionary, here we only use the target templates $ \mathbf{T} \in \mathbb{R}^{d \times n}$ as the dictionary. The basic ordinary least square (OLS) for computing the coefficients is given by
\begin{equation}
\label{eq:OLS}
\hat{\boldsymbol{\alpha}}_{OLS} = \arg\min_{\boldsymbol{\alpha}} \|\mathbf{y} - \mathbf{T}\boldsymbol{\alpha}\|_2^2 
\end{equation}
which has a least square approximation solution
\begin{equation}
\label{eq:OLSsol}
\hat{\boldsymbol{\alpha}}_{OLS} = (\mathbf{T}^{T}\mathbf{T})^{-1}\mathbf{T}^{T}\mathbf{y}
\end{equation}
Often, as in visual tracking, there is a linear dependency between two or more columns of $\mathbf{T}$ which causes to the precision of OLS become very poor. The columns in this case are called \textit{multi-colinear} and may occur in two forms: (1) Exact multi-colinearity: the matrix $\mathbf{T}$ is singular. (2) Near multi-colinearity: at least one of the eigenvalues of the grammian matrices $\mathbf{T}^T\mathbf{T}$ or $\mathbf{T}\mathbf{T}^T$ is very small. In this condition, the linear system obtained becomes ill-conditioned and prohibits us from deriving a reliable linear representation.
\begin{figure}
\begin{center}
\begin{tabular}{cc}
\bmvaHangBox{\fbox{\includegraphics[width=5.6cm]{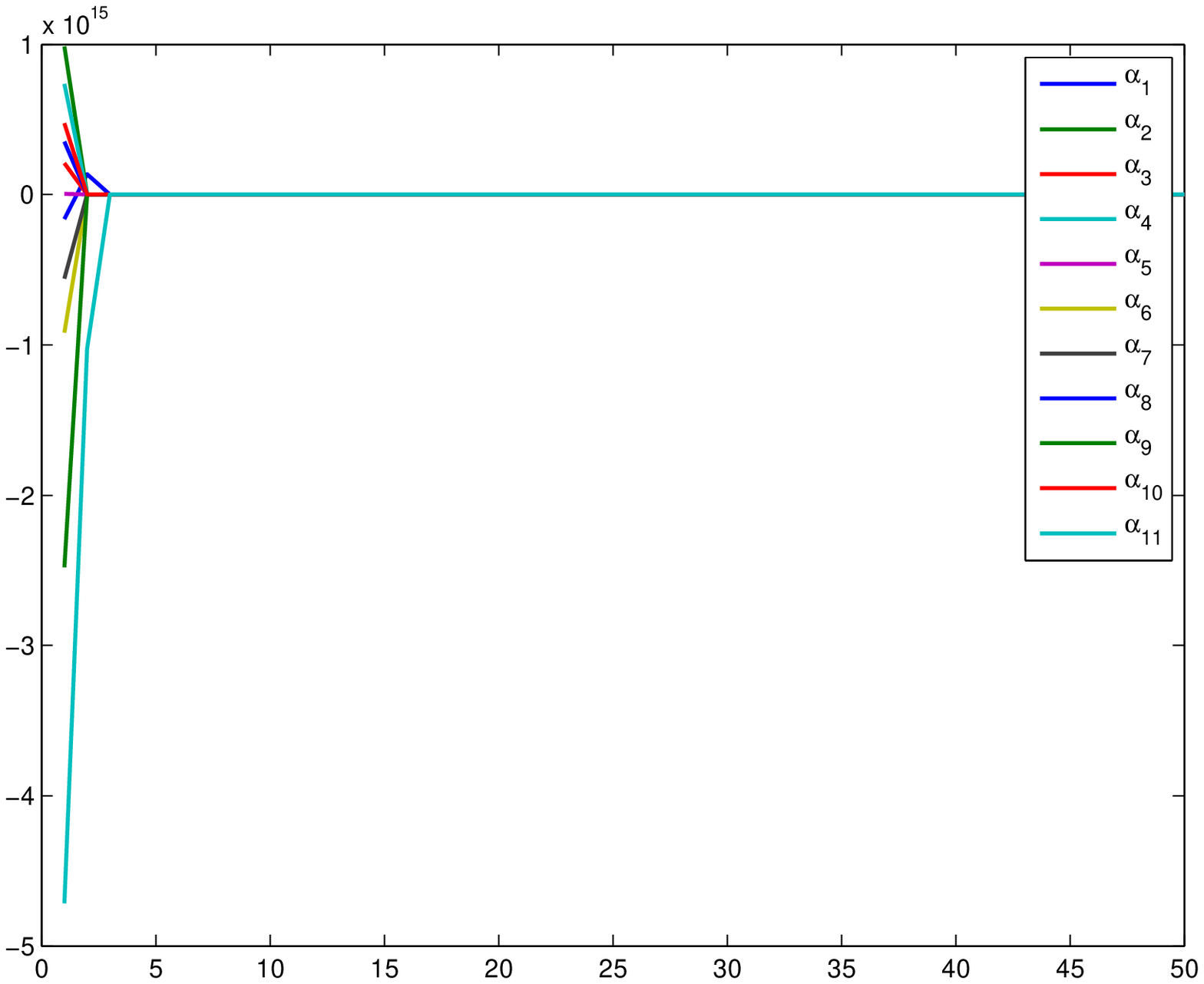}}}& 
\bmvaHangBox{\fbox{\includegraphics[width=5.9cm]{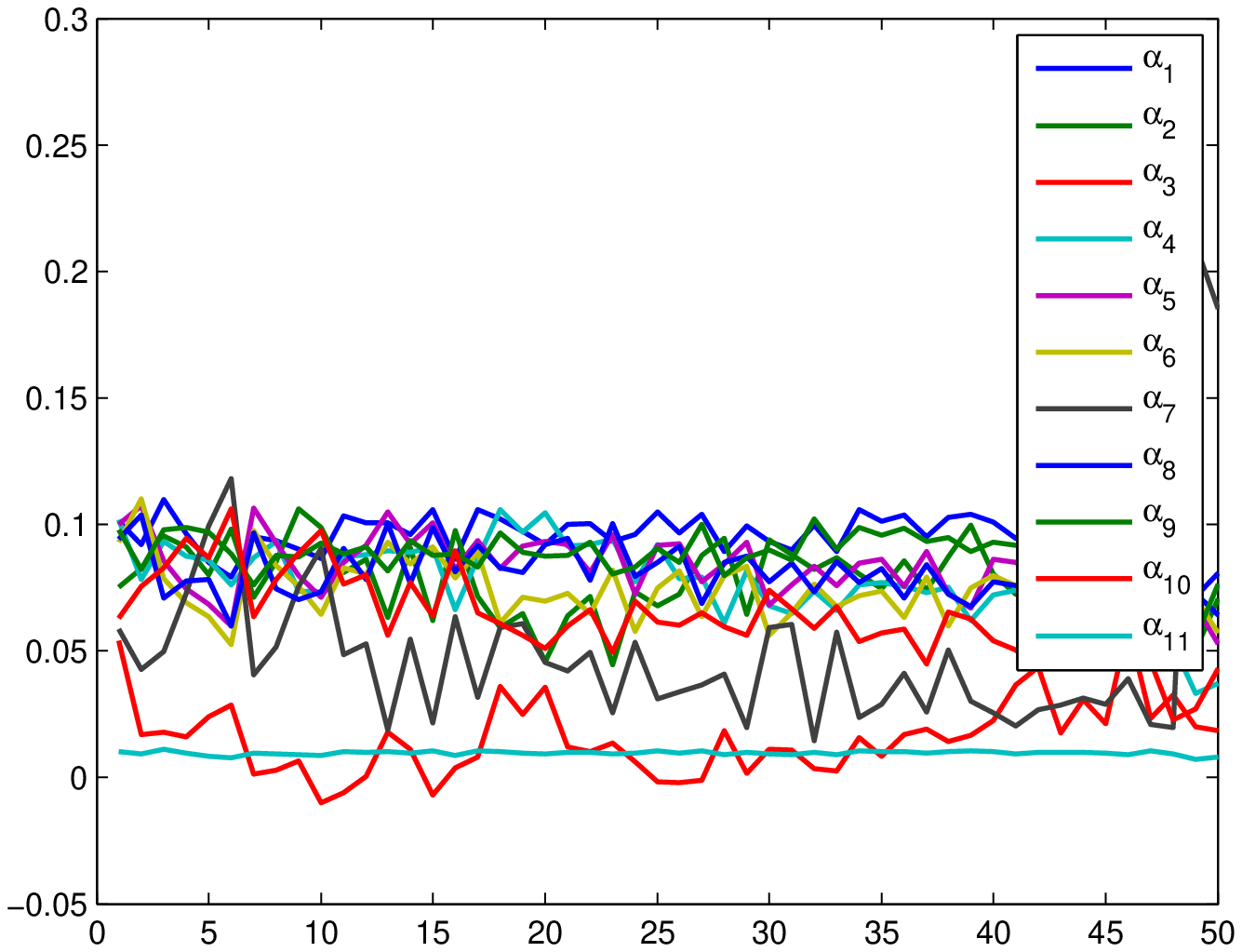}}}\\ 
(a)OLS & (b)RR 
\end{tabular}
\end{center}
\caption{Comparing solution coefficients between OLS and RR models. Heed to the difference of scales for y-axes in two cases which is extremely larger for OLS. After some frames the tracker using OLS drifts and the coefficients estimated become invalid.}
	%As can be seen the existence of occlusion templates leads to uncontrollable activation of them and drift of the tracker eventually. }
\label{fig2}
\end{figure}
In such condition, a reasonable remedy can be obtained through $\ell_{2}$-regularization
\begin{equation}
\label{eq:RR}
\hat{\boldsymbol{\alpha}}_{ridge}  = \argmin\limits_{\boldsymbol{\alpha}}  \left\|\mathbf{y} - \mathbf{T} \boldsymbol{\alpha}\right\|_2^2 + \lambda_{ridge} \left\|\boldsymbol{\alpha}\right\|_2^2
\end{equation}
where $\lambda_{ridge}$ is a constant regulatory parameter that makes a trade-off between the reconstruction error and the energy of coefficients. RR admits a direct analysis solution given by Eq.~\eqref{eq:RRsol} 		
\begin{equation}
\label{eq:RRsol}
\hat{\boldsymbol{\alpha}}_{ridge} = (\mathbf{T}^T\mathbf{T}+\lambda_{ridge}\mathbf{I})^{-1}\mathbf{T}^{T}\mathbf{y}
\end{equation}
where $I \in \mathbb{R}^{n \times n}$ is the identity matrix. In statistics, Eq.~\eqref{eq:RR} is known as \textit{ridge regression} (RR) and was first introduced by Hoerl and Kennard~\cite{hoerl1970ridge}; in vision community it is also known as~\textit{collaborative representation} (CR). To demonstrate the effects of this condition on the estimation of coefficient, we consider sum of coefficients variances (total variance) for $\hat{\boldsymbol{\alpha}}_{OLS}$ and $\hat{\boldsymbol{\alpha}}_{ridge}$ which is given by
\begin{equation}
\label{eq:tvOLS}
TV(\hat{\boldsymbol{\alpha}}_{OLS}) = \sigma^{2}.\sum_{j=1}^{s}\frac{1}{\lambda_j}
\end{equation}
in which $\lambda_j$ is the j-th eigenvalue of $\mathbf{T}$. It can be seen that total variance of OLS would be severely inflated if one or more columns are co-linear. For RR, Eq.~\eqref{eq:tvOLS} becomes
\begin{equation}
\label{eq:tvRR}
TV(\hat{\boldsymbol{\alpha}}_{ridge}) = \sigma^{2}.\sum_{j=1}^{s}\frac{\lambda_j}{(\lambda_j+\lambda_{ridge})^2}
\end{equation}
By comparing Eq.~\eqref{eq:tvOLS} and Eq.~\eqref{eq:tvRR}, it can be noted that for any $\lambda_{ridge} > 0$, RR has a smaller total variance compared to OLS. In Figure~\ref{fig2}, we compare the estimated solution coefficients for a randomly selected video under the OLS and RR which could be seen the coefficients are extremely unstable for OLS (in the range of $10^{15}$) which is by far larger than RR with stabilized coefficients.  As for related works,  Zhang \emph{et al.}~\cite{Zhang11ICCV} showed that great face recognition results reported by~\cite{Wright:2009:RFR:1495801.1496037} were not achieved necessarily on the sparsity constraint and reported competitive results with collaborative representation which replaced $\ell_1$-norm regularization with $\ell_2$-norm regularization in sparse representation model. The advantage of this model was suggested as a simple yet efficient solution compared to sparse representation as the optimization model admits a direct and efficient analytic solution. Li \emph{et al.}~\cite{li2012non} proposed a non-sparse based tracker that used a Mahalanobis distance metric (instead of Euclidean distance) for classification. The drawback of their approaches approach was the estimation of the weight matrix accurately which can be slow for visual tracking for which the authors proposed learning the weight matrix in an online fashion. 
\begin{table}[b]
\footnotesize
\caption{Characteristics of the compared trackers obtained from~\cite{Smeulders2014} and modified. PF: Particle Filter}
\begin{tabular} {|p{1.40cm}|p{1.08cm}|p{1.85cm}|p{1.0cm}|p{2.33cm}|p{2.7cm}|}
\hline
\textbf{Tracker} &\textbf{Dictionary}  & \textbf{Appearance model} & \textbf{Motion model} & \textbf{Optimization method} & \textbf{Update mechanism} \\
\hline
Proposed RR & $\mathbf{D} = \mathbf{T} $ &linear representaion intensities& Gaussian, PF &$\ell_{2}$- regularization &Update bounding boxes, cosine similarity\\
\hline
L1-APG~\cite{bao2012real} & $\mathbf{D} =[\mathbf{T}, \mathbf{I}] $ &linear representaion, intensities& Gaussian, PF &$\ell_{1}$-regularization, constrained particles &Update bounding boxes, cosine similarity\\
\hline
L1-{WMB}~\cite{MeiMinimumCVPR2011} & $\mathbf{D} =[\mathbf{T}, \mathbf{I}] $ &linear representaion,  intensities& Gaussian, PF &$\ell_{1}$-regularization, constrained particles&Update bounding boxes, cosine similarity\\
\hline
L1-{Original}~\cite{Mei09} & $\mathbf{D} =[\mathbf{T}, \mathbf{I}] $ &linear representaion, intensities & Gaussian, PF &$\ell_{1}$-regularization &Update bounding boxes, cosine similarity\\
\hline
\end{tabular}
\label{table:characs}
\end{table}

\section{Experiments}
\label{sec:exp}
\subsection{Experimental setup}
The proposed RR-based tracker (with $\ell_{2}$-norm penalization)  is compared against three state-of-the-art sparse trackers based on $\ell_{1}$-norm penalization including L1-APG~\cite{bao2012real} (Accelerated Proximal Gradient), L1-WMB~\cite{MeiMinimumCVPR2011} (With Minimum Bound) and L1-Original~\cite{Mei09}. Their underlying working characteristics are compared in Table~\ref{table:characs}. The main differences between the RR-based tracker and the compared trackers are in the complexity of the dictionary they use (\emph{i.e.}, $\mathbf{T}$: basis template dictionary versus $\mathbf{D} =[\mathbf{T}, \mathbf{I}]$: augmented dictionary) and in the optimization model they are built upon (\emph{i.e.} $\ell_{1}$ vs. $\ell_{2}$). It is important to point out the following remarks: (1) The sparse $\ell_{1}$ trackers all use an augmented dictionary; (2) Our proposed non-sparse RR-based tracker does not use the occlusion  dictionary because only under the $\ell_{1}$-norm context that promotes sparsity, the use of occlusion  dictionary was suggested to be useful for handling occlusion and such a judgment cannot be made in the $\ell_{2}$-norm context.

We conducted extensive quantitative  experiments on a total of 33 video sequences  which are diverse and contain variety of tracking challenges. These video sequences are collected from the large scale benchmark library presented in~\cite{WuLimYang13} as well as~\cite{babenko11}. For all trackers, we provide quantitative evaluation criterion defined by  the Center Location Error (CLE) and Tracking Success Rate (TSR) which are computed based on a given ground truth. Since the trackers can have a dependency on the random number generation (RNG), we set the seed for the RNG to a fixed non-negative value which would allow us to have a fair comparison between all trackers under similar conditions. Furthermore, in order not be biased to only one realization of random numbers, without reinitialization from the same seed we obtain a sequence of random numbers and run each tracking algorithm 10 times on each video sequence according to the same random number. Results are reported in terms of the average ($\overline{CLE} = \frac{1}{N}.\sum_{i=1}^{N}{CLE_i}~,~\overline{TSR} = \frac{1}{N}.\sum_{i=1}^{N}{TSR_i}$) and standard deviation of the results obtained where $N = 10$ is the number of evaluation and $CLE_i$ and $TSR_i$ are the average CLE and TSR over the entire frames in each run. The parameters related to the particle file variance parameters in our experiment were set to be like in the benchmark~$[0.03 ,0.0005 ,0.0005 ,0.03 ,1 ,1]$ where $\mathbf{t}_x$ = $\mathbf{t}_y$ = $1$ are translations in $\mathbf{x}$ and $\mathbf{y}$ directions. In some videos containing fast motion or pose change (\emph{e.g.} CarDark, Coke, Deer etc.) the variances were changed correspondingly to for example $[0.03 ,0.0005 ,0.0005 ,0.03 ,2 ,2]$ to be able to capture the fast motions, the same for all tracker. The regulatory parameter for $\ell_{1}$ trackers were used as they were used by the original codes. The regulatory parameter for our proposed algorithm was set to~$\lambda_{ridge}=1$ or~$\lambda_{ridge}=2$ in most videos which resulted in fairly similar performance. In rare cases containing severe occlusion (e.g. Coke) ~$\lambda_{ridge}$ was increased to higher values which had a positive effect because it avoided frequent update of the dictionary and insertion of bad template in the dictionary.
\begin{table}[t]
\caption{Comparison of ours vs. three state-of-the-art approaches based on average center location error (CLE). The results in bold are significantly different with an~$\alpha$-confidence level of 5\%}
\begin{center}
\resizebox{\linewidth}{!}{
\begin{tabular} {|l||c||c||c||c||c||c||c||c||c|}
\hline
\textbf{No.} & \textbf{Seq} &\textbf{Proposed RR} &\textbf{rank} & \textbf{L1-APG} &\textbf{rank} & \textbf{L1-WMB} &\textbf{rank} &\textbf{L1-Original} &\textbf{rank} \\
\hline
1 &Car4 &$\textbf{6.78} \pm \textbf{0.27}$ &$1$  &$\textbf{6.59} \pm \textbf{0.40}$ &$1$ &$111.50 \pm 19.98$ &$2$ &$115.90 \pm 17.87$ &$2$\\
2 &CarDark &$\textbf{13.51} \pm \textbf{4.85}$ &$1$ &$\textbf{14.77} \pm \textbf{5.59}$ &$1$ &$36.65 \pm 12.05$ &$2$ &$46.82 \pm 21.27$ &$2$\\
3 &CarScale &$\textbf{12.62} \pm \textbf{2.7}$ &$1$ &$\textbf{15.79} \pm \textbf{1.98}$ &$1$ &$47.37 \pm 24.42$ &$2$ &$62.37 \pm 35.93$ &$2$\\
4 &Cliffbar &$\textbf{3.76} \pm \textbf{2.44}$ &$1$ &$\textbf{5.83 }\pm \textbf{1.36}$ &$1$ &$10.74 \pm 2.62$ &$2$ &$12.11 \pm 2.94$ &$2$\\
5 &Coke &$\textbf{27} \pm \textbf{25.19}$ &$1$ &$123.8 \pm 31.75$ &$2$ &$118.9 \pm 13.75$ &$2$ &$124.6 \pm 11.46$ &$2$\\
6 &Couple &$\textbf{18.87} \pm \textbf{2.55}$ &$1$ &$\textbf{32.67} \pm \textbf{16.11}$  &$1$ &$60.69 \pm 24.76$  &$2$ &$78.66 \pm 24.06$  &$2$\\
7 &Crossing &$\textbf{1.81} \pm \textbf{0.27}$ &$1$ &$\textbf{7.1 }\pm \textbf{16.17}$ &$1$ &$\textbf{2.1} \pm \textbf{0.08}$ &$1$ &$\textbf{2.06} \pm \textbf{0.14}$ &$1$\\
8 &David2 &$\textbf{3.96} \pm \textbf{6.14}$ &$1$ &$\textbf{3.71} \pm \textbf{1.52}$ &$1$ &$64.14 \pm 5.77$ &$2$ &$58.26 \pm 16.39$ &$2$\\
9 &David &$\textbf{8.37} \pm \textbf{2.89}$ &$1$ &$\textbf{11.29} \pm \textbf{5.69}$ &$1$ &$\textbf{7.92} \pm \textbf{3.73}$ &$1$ &$\textbf{7.35} \pm \textbf{4.08}$ &$1$\\
10 &Deer &$\textbf{6.56} \pm \textbf{0.75}$ &$1$ &$\textbf{35.99} \pm \textbf{36.68}$ &$1$ &$100.37 \pm 42.5$ &$2$ &$85.84 \pm 29.04$ &$2$\\
11 &Doll &$\textbf{4.37} \pm \textbf{0.24}$ &$1$ &$\textbf{3.73} \pm \textbf{0.79}$ &$1$ &$48.26 \pm 24.29$ &$2$ &$62.73 \pm 30.19$ &$2$\\
12 &Dollar &$\textbf{2.29 }\pm \textbf{0.65}$ &$1$ &$13.42 \pm 0.25$ &$2$ &$14.41 \pm 3.4$ &$2$ &$13.56 \pm 0.24$ &$2$\\
13 &Dudek &$\textbf{10.92} \pm \textbf{1.54}$ &$1$ &$93.12 \pm 62.75$ &$2$ &$136.1 \pm 52.31$ &$2$ &$101.2 \pm 43.77$ &$2$\\
14 &FaceOcc1 &$\textbf{17.11 }\pm \textbf{2.29}$ &$1$ &$\textbf{15.04} \pm \textbf{0.67}$ &$1$ &$75.62 \pm 59.19$ &$2$ &$\textbf{54.79} \pm \textbf{53.44}$ &$1$\\
15 &FaceOcc2 &$\textbf{12.32} \pm \textbf{3.66}$ &$1$ &$\textbf{13.75} \pm \textbf{1.23}$ &$1$ &$\textbf{29.74} \pm \textbf{36.29}$ &$1$ &$\textbf{29.9} \pm \textbf{36.74}$ &$1$\\
16 &Fish &$\textbf{43.21} \pm \textbf{16.87}$  &$1$ &$\textbf{41.44} \pm \textbf{22.61}$ &$1$ &$71.1 \pm 29.66$ &$2$ &$\textbf{58.77} \pm \textbf{25.66}$ &$1$\\
17 &FleetFace &$\textbf{20.15} \pm \textbf{2.97}$ &$1$ &$41.33 \pm 21.68$ &$2$ &$114.88 \pm 6.53$ &$3$ &$114.99 \pm 11.4$ &$3$\\
18 &Football1 &$\textbf{17.81} \pm \textbf{9.52}$ &$1$ &$\textbf{28.01} \pm \textbf{13.86}$ &$1$ &$61.35 \pm 9.65$ &$2$ &$58.45 \pm 9.91$ &$2$\\
19 &Football &$\textbf{13.81} \pm \textbf{1.07}$ &$1$ &$60.86 \pm 54.28$ &$2$ &$96.35 \pm 20.72$ &$3$ &$112.5 \pm 11.44$ &$3$\\
20 &Freeman1 &$\textbf{61.03} \pm \textbf{39.38}$ &$1$ &$\textbf{45.32} \pm \textbf{33.13}$ &$1$ &$\textbf{32.21} \pm \textbf{29.98}$ &$1$ &$\textbf{55.8} \pm \textbf{23.9}$ &$1$\\
21 &Freeman3 &$\textbf{7.28} \pm \textbf{4.03}$ &$1$ &$26.69 \pm 10.97$ &$2$ &$55.22 \pm 2.73$ &$2$ &$38.53 \pm 22.28$ &$3$\\
22 &Freeman4 &$\textbf{22.66} \pm \textbf{18.07}$ &$1$ &$\textbf{30.18} \pm \textbf{18.7}$ &$1$ &$57.97 \pm 14.67$ &$2$ &$60.13 \pm 14.76$ &$2$\\
23 &Girl &$\textbf{5.29} \pm \textbf{3.05}$ &$1$ &$\textbf{7.39} \pm \textbf{3.04}$ &$1$ &$\textbf{16.58} \pm \textbf{14.63}$ &$1$ &$\textbf{13.94} \pm \textbf{13.97}$ &$1$\\
24 &Jumping &$\textbf{16.47} \pm \textbf{22.75}$ &$1$ &$\textbf{4.61} \pm \textbf{0.6}$ &$1$ &$\textbf{27.58} \pm \textbf{41.49}$ &$1$ &$\textbf{34.74} \pm \textbf{36.81}$ &$1$\\
25 &Mhyang &$\textbf{2.62} \pm \textbf{0.73}$ &$1$ &$\textbf{3.01} \pm \textbf{1.23}$ &$1$ &$36.27 \pm 6.14$ &$2$ &$38.32 \pm 4.51$ &$2$\\
26 &Mountain Bike &$\textbf{8.16} \pm \textbf{1.98}$ &$1$ &$160.3 \pm 78.48$ &$2$ &$201.5 \pm 57.65$ &$2$ &$207.6 \pm 59.65$ &$2$\\
27 &Singer1 &$\textbf{4.34} \pm \textbf{0.74}$ &$1$ &$\textbf{4.65} \pm \textbf{0.75}$ &$1$ &$\textbf{54.39} \pm \textbf{80.43}$ &$1$ &$\textbf{21.86} \pm \textbf{53.92}$ &$1$\\
28 &Soccer &$\textbf{60.83} \pm \textbf{14.44}$ &$1$ &$\textbf{88.56} \pm \textbf{36.8}$ &$1$ &$152.6 \pm 32.32$ &$2$ &$133.5 \pm 27.75$ &$2$\\
29 &Subway &$36.32 \pm 1.06$ &$2$ &$37.06 \pm 1.49$ &$2$ &$\textbf{3.97} \pm \textbf{0.17}$ &$1$ &$\textbf{4.01} \pm \textbf{0.22}$ &$1$\\
30 &Surfer &$\textbf{1.9} \pm \textbf{0.67}$ &$1$ &$13.38 \pm 0.24$  &$2$ &$13.59 \pm 0.42$  &$2$ &$14.6 \pm 3.84$  &$2$\\
31 &SUV &$\textbf{26.38} \pm \textbf{28.43}$ &$1$ &$51.15 \pm 23.62$ &$2$ &$\textbf{26.17} \pm \textbf{2.13}$ &$1$ &$\textbf{26.46} \pm \textbf{2.53}$ &$1$\\
32 &Sylvester &$\textbf{18.51} \pm \textbf{9.94}$ &$1$ &$\textbf{32.66} \pm \textbf{11.49}$ &$1$ &$\textbf{44.85} \pm \textbf{28.54}$ &$1$ &$51.12 \pm 31.26$ &$2$\\
33 &Trellis &$\textbf{12.95} \pm \textbf{8.15}$ &$1$ &$\textbf{31.66} \pm \textbf{7.45}$ &$1$ &$65.71 \pm 22.82$ &$2$ &$81.03 \pm 34.29$ &$2$\\
\hline
\multicolumn{2}{|c||}{Avg\textbf{ $\overline{CLE}$}/rank} &$\textbf{16.06}$ &$\textbf{1.03}$ &$33.48$ &$1.30$ &$60.51$ &$1.76$ &$60.08$ &$1.76$ \\
\hline
\end{tabular}}
\end{center}
\label{table:CLE}
\end{table}
\subsection{Significance Testing for Quantitative Evaluation}
Table~\ref{table:CLE} and Table~\ref{table:TSR} present the computed $\overline{CLE}$ and $\overline{TSR}$ for the all compared trackers on each test sequence. A multiple pairwise comparison testing based on one-way analysis of variance (ANOVA) is applied on each video sequence to evaluate whether or not the difference between the groups' averages for each tracker most likely reflects a significant difference or not. ANOVA is a generalized significance t-test which is applicable when the test statistic would follow a normal distribution. We argue that the normality assumption can be made on the groups' distributions based on \emph{the central limit theorem}. The reason is e.g. for $CLE_i$ (the i-th evaluation of CLE), it is the average of many random variables (errors at each pixel) which can be assumed to be independent and identically distributed (i.i.d). Central limit theorem therefore states that the mean of these i.i.d random variables (\emph{i.e.} $CLE_i$) follows a normal distribution and therefore a significance t-test is applicable on the group of $CLE_i$'s obtained from different evaluations. The result of such ANOVA-based significance test is provided in Tables~\ref{table:CLE} and Table~\ref{table:TSR} as a ranking value on each video which implies based on $\alpha$-level significance test ($\alpha = 5\%$), if two algorithms have significantly different performance or not.
\begin{table}[t]
%\footnotesize
\caption{Comparison of ours vs. three state-of-the-art approaches based on tracking success rate (TSR). The results in bold are significantly different with an~$\alpha$-confidence level of 5\%.}
\begin{center}
\resizebox{\linewidth}{!}{
\begin{tabular} {|l||c||c||c||c||c||c||c||c||c|}%{p{0.15cm} p{0.9cm} p{1.58cm} p{0.28cm} p{1.55cm} p{0.28cm} p{1.69cm} p{0.28cm} p{1.69cm} p{0.28cm}}
\hline
\textbf{No.} & \textbf{Seq} &\textbf{Proposed RR} &\textbf{rank} & \textbf{L1-APG} &\textbf{rank} & \textbf{L1-WMB} &\textbf{rank} &\textbf{L1-Original} &\textbf{rank} \\
\hline
1 &Car4 &$\textbf{1} \pm \textbf{0}$ &$1$ &$\textbf{1} \pm \textbf{0}$ &$1$ &$0.14 \pm 0.14$ &$2$ &$0.09 \pm 0.13$ &$2$\\
2 &CarDark &$\textbf{0.74} \pm \textbf{0.09}$ &$1$ &$\textbf{0.72 }\pm \textbf{0.1}$ &$1$ &$\textbf{0.52} \pm \textbf{0.18}$ &$1$ &$0.34 \pm 0.29$ &$2$\\
3 &CarScale &$\textbf{0.83 }\pm \textbf{0.07}$ &$1$ &$\textbf{0.79} \pm \textbf{0.01}$ &$1$ &$0.6 \pm 0.2$ &$2$ &$0.47 \pm 0.3$ &$2$\\
4 &Cliffbar &$\textbf{0.43} \pm \textbf{0.1}$ &$1$ &$\textbf{0.4} \pm \textbf{0.05}$ &$1$ &$0.3 \pm 0.13$ &$2$ &$0.24 \pm 0.13$ &$2$\\
5 &Coke &$\textbf{0.67} \pm \textbf{0.25}$ &$1$ &$0.06 \pm 0.01$ &$2$ &$0.03 \pm 0.01$ &$2$ &$0.02 \pm 0.01$ &$2$\\
6 &Couple &$\textbf{0.52 }\pm \textbf{0.07}$ &$1$ &$\textbf{0.44} \pm \textbf{0.09}$ &$1$ &$0.21 \pm 0.14$ &$2$ &$0.11 \pm 0.15$ &$2$\\
7 &Crossing &$\textbf{0.95} \pm \textbf{0.04}$ &$1$ &$\textbf{0.88} \pm \textbf{0.23}$ &$1$ &$\textbf{0.98} \pm \textbf{0.04}$ &$1$ &$\textbf{0.97} \pm \textbf{0.02}$ &$1$\\
8 &David2 &$\textbf{0.93} \pm \textbf{0.14}$ &$1$ &$\textbf{0.83} \pm \textbf{0.11}$ &$1$ &$0.15 \pm 0.08$ &$2$ &$0.23 \pm 0.14$ &$2$\\
9 &David &$\textbf{0.22} \pm \textbf{0.01}$ &$1$ &$\textbf{0.22} \pm \textbf{0.01}$ &$1$ &$\textbf{0.21} \pm \textbf{0.01}$ &$1$ &$\textbf{0.22} \pm \textbf{0.02}$ &$1$\\
10 &Deer &$\textbf{0.99} \pm \textbf{0.02}$ &$1$ &$\textbf{0.78} \pm \textbf{0.24}$ &$1$ &$0.39 \pm 0.28$ &$2$ &$0.44 \pm 0.28$ &$2$\\
11 &Doll &$0.51 \pm 0.17$ &$2$ &$\textbf{0.81} \pm \textbf{0.16}$ &$1$ &$0.54 \pm 0.14$ &$2$ &$0.42 \pm 0.19$ &$2$\\
12 &Dollar &$\textbf{0.92} \pm \textbf{0.17}$ &$1$ &$0.39 \pm 0$ &$2$ &$0.35 \pm 0.12$ &$2$ &$0.37 \pm 0.02$ &$2$\\
13 &Dudek &$\textbf{0.78 }\pm \textbf{0.03}$ &$1$ &$\textbf{0.68} \pm \textbf{0.09}$ &$1$ &$0.31 \pm 0.31$ &$2$ &$0.51 \pm 0.26$ &$2$\\
14 &FaceOcc1 &$\textbf{0.91} \pm \textbf{0.1}$ &$1$ &$\textbf{0.98} \pm \textbf{0.02}$ &$1$ &$0.45 \pm 0.46$ &$2$ &$\textbf{0.6} \pm \textbf{0.41}$ &$1$\\
15 &FaceOcc2 &$\textbf{0.42} \pm \textbf{0.08}$ &$1$ &$\textbf{0.38} \pm \textbf{0.03}$ &$1$ &$0.27 \pm 0.14$ &$2$ &$\textbf{0.3} \pm \textbf{0.15}$ &$1$\\
16 &Fish &$\textbf{0.05 }\pm \textbf{0.01}$ &$1$ &$\textbf{0.14} \pm \textbf{0.14}$ &$1$ &$\textbf{0.07} \pm \textbf{0.06}$ &$1$ &$\textbf{0.09 }\pm \textbf{0.06}$ &$1$\\
17 &FleetFace &$\textbf{0.64} \pm \textbf{0.01}$ &$1$ &$\textbf{0.64} \pm \textbf{0.02}$ &$1$ &$0.52 \pm 0.02$ &$2$ &$0.52 \pm 0.01$&$2$ \\
18 &Football1 &$\textbf{0.57 }\pm \textbf{0.16}$ &$1$ &$0.32 \pm 0.1$ &$2$ &$0.12 \pm 0.05$ &$3$ &$0.15 \pm 0.08$ &$3$\\
19 &Football &$\textbf{0.7} \pm \textbf{0.06}$ &$1$ &$0.45 \pm 0.12$ &$2$ &$0.14 \pm 0.07$ &$3$ &$0.12 \pm 0.08$ &$3$\\
20 &Freeman1 &$\textbf{0.19 }\pm \textbf{0.04}$ &$1$ &$\textbf{0.17 }\pm \textbf{0.05}$ &$1$ &$\textbf{0.23 }\pm \textbf{0.11}$ &$1$ &$\textbf{0.22} \pm \textbf{0.04}$ &$1$\\
21 &Freeman3 &$\textbf{0.71} \pm \textbf{0.15}$ &$1$ &$0.59 \pm 0.09$ &$2$ &$0.59 \pm 0.01$ &$2$ &$\textbf{0.6} \pm \textbf{0.06}$ &$1$\\
22 &Freeman4 &$\textbf{0.3} \pm \textbf{0.1}$ &$1$ &$\textbf{0.35} \pm \textbf{0.11}$ &$1$ &$0.24 \pm 0.02$ &$2$ &$0.24 \pm 0.04$ &$2$\\
23 &Girl &$\textbf{0.67} \pm \textbf{ 0.15}$ &$1$ &$\textbf{0.46} \pm \textbf{0.16}$ &$1$ &$\textbf{0.48} \pm \textbf{0.37}$ &$1$ &$\textbf{0.54 }\pm \textbf{0.35}$ &$1$\\
24 &Jumping &$\textbf{0.71} \pm \textbf{0.25}$ &$1$ &$\textbf{0.94} \pm \textbf{0.06}$ &$1$ &$\textbf{0.7} \pm \textbf{0.37}$ &$1$ &$0.55 \pm 0.38$ &$2$\\
25 &Mhyang &$\textbf{0.99} \pm \textbf{0.03}$ &$1$ &$\textbf{0.98} \pm \textbf{0.03}$ &$1$ &$0.6 \pm 0.09$ &$2$ &$0.54 \pm 0.12$ &$2$\\
26 &Mountain Bike &$\textbf{0.74} \pm \textbf{0.16}$ &$1$ &$0.37 \pm 0.12$ &$2$ &$0.06 \pm 0.05$ &$3$ &$0.06 \pm 0.05$ &$3$\\
27 &Singer1 &$\textbf{0.97} \pm \textbf{0.11}$ &$1$ &$0.96 \pm 0.07$ &$2$ &$0.69 \pm 0.47$ &$2$ &$0.85 \pm 0.3$ &$2$\\
28 &Soccer &$\textbf{0.18 }\pm \textbf{0.02}$ &$1$ &$\textbf{0.15} \pm \textbf{0.02}$ &$1$ &$0.05 \pm 0.04$ &$2$ &$0.08 \pm 0.03$ &$2$\\
29 &Subway &$\textbf{0.5} \pm \textbf{0.02}$ &$1$ &$\textbf{0.5} \pm \textbf{0.01}$ &$1$ &$0.89 \pm 0.03$ &$2$ &$0.88 \pm 0.03$ &$2$\\
30 &Surfer &$\textbf{1} \pm \textbf{0}$ &$1$ &$0.39 \pm 0$ &$2$ &$0.38 \pm 0.02$ &$2$ &$0.35 \pm 0.12$ &$2$\\
31 &SUV &$\textbf{0.85} \pm \textbf{0.16}$ &$1$ &$0.65 \pm 0.12$ &$2$ &$0.69 \pm 0$ &$2$ &$0.69 \pm 0$ &$2$\\
32 &Sylvester &$\textbf{0.65} \pm \textbf{0.12}$ &$1$ &$0.43 \pm 0.06$ &$2$ &$0.24 \pm 0.12$ &$3$ &$0.21 \pm 0.14$ &$3$\\
33 &Trellis &$\textbf{0.51} \pm \textbf{0.13}$ &$1$ &$0.32 \pm 0.09$ &$2$ &$0.19 \pm 0.12$ &$2$ &$0.18 \pm 0.13$ &$2$\\
\hline
\multicolumn{2}{|c||}{Avg\textbf{ $\overline{TSR}$}/rank} &$\textbf{0.66}$ &$\textbf{1.03}$ &$0.55 $ &$1.33$ &$0.37$ &$1.82$ &$0.37$ &$1.88$\\
\hline
 \end{tabular} }
\end{center}
\label{table:TSR}
\end{table}
For example in the video Deer, the computed $\overline{CLE}$ for RR and L1-APG are $6.56 \pm 0.75$ and $35.99 \pm 36.68$ respectively. While the absolute value of the $\overline{CLE}$'s are greatly different, their $\alpha$-level significance test show that they are not significantly different and thus they are both given the same ranking equal to $1$. 
\begin{figure}[t]
\caption{Comparing the performance of two competing trackers in handling different challenging tracking scenarios during the tracking process}
\begin{center}
\begin{tabular}{c}
\bmvaHangBox{\fbox{\includegraphics[width=0.61\textwidth]{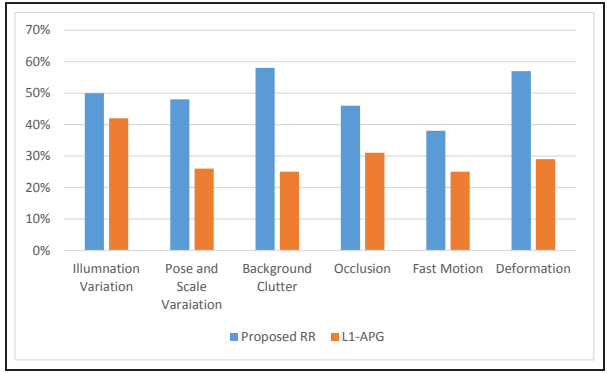}}} 
\end{tabular}
\end{center}
\label{fig:challenges}
\end{figure}
In another video Dollar for instance, the $\overline{CLE}$ of RR and L1-APG with $2.29 \pm 0.65$ and $13.42 \pm 0.25$ based on the significance test are considered significantly different. In this manner, we are able to conduct a fair comparison between the performance of trackers based on the results obtained from different evaluations.
\subsection{Comparison of competing trackers}
The performance of the proposed RR-based tracker against the competing $\ell_{1}$ trackers can be compared in Table~\ref{table:CLE} and Table~\ref{table:TSR} according to the average rankings computed by averaging out the rankings computed based on $\alpha$-level significance test on each video. As could be seen our proposed RR-based tracker has the best ranking (\emph{i.e.} $1.03$ and $1.03$) against the competing trackers which shows it is capable of effectively handling complicated appearance changes in the tracking process. In Figure~\ref{fig:challenges}, we also provide the performance of the two best competing trackers under different tracking challenges as a means to compare their performances under such circumstances. The vertical axis is the percentage of videos with a particular challenge for which RR and L1-APG trackers pass it successfully. The challenges for each video were obtained found~\cite{WuLimYang13}.As could be seen, RR outperforms L1-APG almost in all challenges. It could be as noted that both trackers are weak in handling fast motions which is the drawback of these trackers. Finally, the efficiency of the proposed tracker against the competing trackers  in terms of average speed is compared in Table~\ref{table:FPS} and the results are greatly in favor of the proposed RR-based tracker.
\begin{table}[h]
\footnotesize
\caption{Comparison of the proposed vs. three state-of-the-art approaches based on average running speed in terms of frames/sec. The first best result is labeled by bold.}
\begin{center}
\begin{tabular}{p{1.5cm} p{1.9cm} p{1.55cm} p{1.69cm} p{1.69cm}}
\hline
\textbf{Seq} &\textbf{Proposed RR} & \textbf{L1-APG}  & \textbf{L1-WMB}  &\textbf{L1-Original}  \\
\hline
\textbf{Avg Speed} &$\textbf{10.34}$ &$4.85 $ &$3.09$ &$3.1$\\
\hline
 \end{tabular}
\end{center}
\label{table:FPS}
\end{table}
\section{Conclusion}
\label{sec:con}
Before the stress on sparsity and using complex dictionaries for handling occlusions etc., we have shown in this paper that the main problem in visual tracking arises from colinearity of data which could be solved by classical ridge regression. Indeed, too much push on sparsity leads to penalization of results with respects to classical ridge regression. To this end, a robust visual tracker based on non-sparse linear representation was proposed that can effectively handle different tracking challenges in extended tracking sequences. The results indicate that our proposed tracker can archive competitively better results compared to $\ell_{1}$ trackers while having faster running speed, which supports the effectiveness of our proposed non-sparse tracker for practical applications.    
\bibliography{egbib}

\begin{thebibliography}{30}
\providecommand{\natexlab}[1]{#1}
\providecommand{\url}[1]{\texttt{#1}}
\expandafter\ifx\csname urlstyle\endcsname\relax
  \providecommand{\doi}[1]{doi: #1}\else
  \providecommand{\doi}{doi: \begingroup \urlstyle{rm}\Url}\fi

\bibitem[Al-rahayfeh and Member(2013)]{Al-rahayfeh2013}
Amer Al-rahayfeh and Miad~Faezipour Member.
\newblock {Eye Tracking and Head Movement Detection : A State-of-Art Survey}.
\newblock \emph{IEEE Journal of Translational Engineering in Health and
  Medicine}, \penalty0 (August), 2013.

\bibitem[Atev et~al.(2005)Atev, Arumugam, Masoud, Janardan, and
  Papanikolopoulos]{Atev:2005:VAC:2218579.2218774}
S.~Atev, H.~Arumugam, O.~Masoud, R.~Janardan, and N.~P. Papanikolopoulos.
\newblock A vision-based approach to collision prediction at traffic
  intersections.
\newblock \emph{Trans. Intell. Transport. Sys.}, 6\penalty0 (4):\penalty0
  416--423, December 2005.
\newblock ISSN 1524-9050.
\newblock \doi{10.1109/TITS.2005.858786}.
\newblock URL \url{http://dx.doi.org/10.1109/TITS.2005.858786}.

\bibitem[Babenko and Belongie(2011)]{babenko11}
Boris Babenko and Ming-Hsuan Yang~Serge Belongie.
\newblock Robust object tracking with online multiple instance learning.
\newblock 2011.

\bibitem[Bao et~al.(2012)Bao, Wu, Ling, and Ji]{bao2012real}
Chenglong Bao, Yi~Wu, Haibin Ling, and Hui Ji.
\newblock Real time robust l1 tracker using accelerated proximal gradient
  approach.
\newblock In \emph{Computer Vision and Pattern Recognition (CVPR), 2012 IEEE
  Conference on}, pages 1830--1837. IEEE, 2012.

\bibitem[Coifman et~al.(1998)Coifman, Beymer, Mclauchlan, and
  Malik]{Coifman1998}
Benjamin Coifman, David Beymer, Philip Mclauchlan, and Jitendra Malik.
\newblock {A real-time computer vision system for vehicle tracking and
  surveillance}.
\newblock \emph{Transportation Research Part C}, 6:\penalty0 271--288, 1998.

\bibitem[Deldjoo()]{deldjoo2009wii}
Yashar Deldjoo.
\newblock \emph{Wii remote based head tracking in 3D audio rendering}.

\bibitem[Deldjoo and Atani(2016)]{deldjoo2016low}
Yashar Deldjoo and Reza~Ebrahimi Atani.
\newblock A low-cost infrared-optical head tracking solution for virtual 3d
  audio environment using the nintendo wii-remote.
\newblock \emph{Entertainment Computing}, 12:\penalty0 9--27, 2016.

\bibitem[Hariharakrishnan and
  Schonfeld(2005)]{Hariharakrishnan:2005:FOT:2219086.2219477}
K.~Hariharakrishnan and D.~Schonfeld.
\newblock Fast object tracking using adaptive block matching.
\newblock \emph{Trans. Multi.}, 7\penalty0 (5):\penalty0 853--859, October
  2005.
\newblock ISSN 1520-9210.
\newblock \doi{10.1109/TMM.2005.854437}.
\newblock URL \url{http://dx.doi.org/10.1109/TMM.2005.854437}.

\bibitem[Hoerl and Kennard(1970)]{hoerl1970ridge}
Arthur~E Hoerl and Robert~W Kennard.
\newblock Ridge regression: Biased estimation for nonorthogonal problems.
\newblock \emph{Technometrics}, 12\penalty0 (1):\penalty0 55--67, 1970.

\bibitem[Hu et~al.(2004)Hu, Tan, Wang, and
  Maybank]{Hu:2004:SVS:2220414.2220805}
Weiming Hu, Tieniu Tan, Liang Wang, and S.~Maybank.
\newblock A survey on visual surveillance of object motion and behaviors.
\newblock \emph{Trans. Sys. Man Cyber Part C}, 34\penalty0 (3):\penalty0
  334--352, August 2004.
\newblock ISSN 1094-6977.
\newblock \doi{10.1109/TSMCC.2004.829274}.
\newblock URL \url{http://dx.doi.org/10.1109/TSMCC.2004.829274}.

\bibitem[Ji and Wang(2015)]{ji2015object}
Zhangjian Ji and Weiqiang Wang.
\newblock Object tracking based on local dynamic sparse model.
\newblock \emph{Journal of Visual Communication and Image Representation},
  2015.

\bibitem[Kastrinaki et~al.(2003)Kastrinaki, Zervakis, and
  Kalaitzakis]{Kastrinaki2003}
V~Kastrinaki, M~Zervakis, and K~Kalaitzakis.
\newblock A survey of video processing techniques for traffic applications.
\newblock \emph{Image and Vision Computing}, 21\penalty0 (4):\penalty0
  359--381, April 2003.
\newblock ISSN 02628856.
\newblock \doi{10.1016/S0262-8856(03)00004-0}.
\newblock URL
  \url{http://linkinghub.elsevier.com/retrieve/pii/S0262885603000040}.

\bibitem[Kim et~al.(2010)Kim, Choi, Yi, Choi, and Kong]{Kim2010}
In~Su Kim, Hong~Seok Choi, Kwang~Moo Yi, Jin~Young Choi, and Seong~G. Kong.
\newblock Intelligent visual surveillance: A survey.
\newblock \emph{International Journal of Control, Automation and Systems},
  8\penalty0 (5):\penalty0 926--939, October 2010.
\newblock ISSN 1598-6446.
\newblock \doi{10.1007/s12555-010-0501-4}.
\newblock URL \url{http://link.springer.com/10.1007/s12555-010-0501-4}.

\bibitem[Li et~al.(2011)Li, Shen, and Shi]{Li11CVPR}
Hanxi Li, Chunhua Shen, and Qinfeng Shi.
\newblock Real-time visual tracking using compressive sensing.
\newblock \emph{Proceedings of the IEEE Conference on Computer Vision and
  Pattern Recognition}, pages 1305 --1312, 2011.

\bibitem[Li et~al.(2012)Li, Shen, Shi, Dick, and van~den Hengel]{li2012non}
Xi~Li, Chunhua Shen, Qinfeng Shi, Anthony Dick, and Anton van~den Hengel.
\newblock Non-sparse linear representations for visual tracking with online
  reservoir metric learning.
\newblock In \emph{Computer Vision and Pattern Recognition (CVPR), 2012 IEEE
  Conference on}, pages 1760--1767. IEEE, 2012.

\bibitem[Mei and Ling(2009)]{Mei09}
X.~Mei and H.~Ling.
\newblock Robust visual tracking using {L}1 minimization.
\newblock \emph{Proceedings of the 12th International Conference on Computer
  Vision}, pages 1436--1443, 2009.

\bibitem[Mei et~al.(2011)Mei, Ling, Wu, Blasch, and Bai]{MeiMinimumCVPR2011}
Xue Mei, Haibin Ling, Yi~Wu, Erik Blasch, and Li~Bai.
\newblock Minimum error bounded efficient {L}1 tracker with occlusion
  detection.
\newblock \emph{Proceedings of the IEEE Conference on Computer Vision and
  Pattern Recognition}, pages 1257--1264, 2011.

\bibitem[Mitchell et~al.(1996)Mitchell, Pennebaker, Fogg, and
  Legall]{Mitchell:1996:MVC:548218}
Joan~L. Mitchell, William~B. Pennebaker, Chad~E. Fogg, and Didier~J. Legall,
  editors.
\newblock \emph{MPEG Video Compression Standard}.
\newblock Chapman \& Hall, Ltd., London, UK, UK, 1996.
\newblock ISBN 0412087715.

\bibitem[Olshausen et~al.(1996)]{olshausen1996emergence}
Bruno~A Olshausen et~al.
\newblock Emergence of simple-cell receptive field properties by learning a
  sparse code for natural images.
\newblock \emph{Nature}, 381\penalty0 (6583):\penalty0 607--609, 1996.

\bibitem[Pavlovic et~al.(1997)Pavlovic, Sharma, and
  Huang]{Pavlovic:1997:VIH:261506.272696}
Vladimir~I. Pavlovic, Rajeev Sharma, and Thomas~S. Huang.
\newblock Visual interpretation of hand gestures for human-computer
  interaction: A review.
\newblock \emph{IEEE Trans. Pattern Anal. Mach. Intell.}, 19\penalty0
  (7):\penalty0 677--695, July 1997.
\newblock ISSN 0162-8828.
\newblock \doi{10.1109/34.598226}.
\newblock URL \url{http://dx.doi.org/10.1109/34.598226}.

\bibitem[Rigamonti et~al.(2011)Rigamonti, Brown, and
  Lepetit]{RigamontiCVPR2011}
R.~Rigamonti, M.A. Brown, and V.~Lepetit.
\newblock Are sparse representations really relevant for image classification?
\newblock \emph{Proceedings of the IEEE Conference on Computer Vision and
  Pattern Recognition}, pages 1545--1552, 2011.

\bibitem[Sikora(1997)]{Sikora:1997:MVS:2322483.2322735}
T.~Sikora.
\newblock The mpeg-4 video standard verification model.
\newblock \emph{IEEE Trans. Cir. and Sys. for Video Technol.}, 7\penalty0
  (1):\penalty0 19--31, February 1997.
\newblock ISSN 1051-8215.
\newblock \doi{10.1109/76.554415}.
\newblock URL \url{http://dx.doi.org/10.1109/76.554415}.

\bibitem[Smeulders et~al.(2014)Smeulders, Member, Chu, Member, Cucchiara, and
  Calderara]{Smeulders2014}
Arnold W~M Smeulders, Senior Member, Dung~M Chu, Student Member, Rita
  Cucchiara, and Simone Calderara.
\newblock {Visual Tracking : An Experimental Survey}.
\newblock 36\penalty0 (7):\penalty0 1442--1468, 2014.

\bibitem[Wright et~al.(2009)Wright, Yang, Ganesh, Sastry, and
  Ma]{Wright:2009:RFR:1495801.1496037}
John Wright, Allen~Y. Yang, Arvind Ganesh, S.~Shankar Sastry, and Yi~Ma.
\newblock Robust face recognition via sparse representation.
\newblock \emph{IEEE Trans. Pattern Anal. Mach. Intell.}, 31\penalty0
  (2):\penalty0 210--227, February 2009.
\newblock ISSN 0162-8828.
\newblock \doi{10.1109/TPAMI.2008.79}.
\newblock URL \url{http://dx.doi.org/10.1109/TPAMI.2008.79}.

\bibitem[Wu et~al.(2013)Wu, Lim, and Yang]{WuLimYang13}
Yi~Wu, Jongwoo Lim, and Ming-Hsuan Yang.
\newblock Online object tracking: A benchmark.
\newblock In \emph{IEEE Conference on Computer Vision and Pattern Recognition
  (CVPR)}, 2013.

\bibitem[Zhang et~al.(2012{\natexlab{a}})Zhang, Zhang, and Yang]{zhang2012real}
Kaihua Zhang, Lei Zhang, and Ming-Hsuan Yang.
\newblock Real-time compressive tracking.
\newblock pages 864--877, 2012{\natexlab{a}}.

\bibitem[Zhang et~al.(2011)Zhang, Yang, and X.]{Zhang11ICCV}
L.~Zhang, M.~Yang, and Feng X.
\newblock Sparse representation or collaborative representation: Which helps
  face recognition?
\newblock \emph{International Conference on Computer Vision}, pages 471--478,
  2011.

\bibitem[Zhang et~al.(2013)Zhang, Yao, Sun, and
  Lu]{Zhang:2013:SCB:2445640.2446001}
Shengping Zhang, Hongxun Yao, Xin Sun, and Xiusheng Lu.
\newblock Sparse coding based visual tracking: Review and experimental
  comparison.
\newblock \emph{Pattern Recogn.}, 46\penalty0 (7):\penalty0 1772--1788, July
  2013.
\newblock ISSN 0031-3203.
\newblock \doi{10.1016/j.patcog.2012.10.006}.
\newblock URL \url{http://dx.doi.org/10.1016/j.patcog.2012.10.006}.

\bibitem[Zhang et~al.(2012{\natexlab{b}})Zhang, Ghanem, Liu, and
  Ahuja]{zhang2012robust}
Tianzhu Zhang, Bernard Ghanem, Si~Liu, and Narendra Ahuja.
\newblock Robust visual tracking via multi-task sparse learning.
\newblock In \emph{Computer Vision and Pattern Recognition (CVPR), 2012 IEEE
  Conference on}, pages 2042--2049. IEEE, 2012{\natexlab{b}}.

\bibitem[Zhong et~al.(2012)Zhong, Lu, and Yang]{zhong2012robust}
Wei Zhong, Huchuan Lu, and Ming-Hsuan Yang.
\newblock Robust object tracking via sparsity-based collaborative model.
\newblock In \emph{Computer vision and pattern recognition (CVPR), 2012 IEEE
  Conference on}, pages 1838--1845. IEEE, 2012.

\end{thebibliography}


\begin{thebibliography}{6}
\providecommand{\natexlab}[1]{#1}
\providecommand{\url}[1]{\texttt{#1}}
\expandafter\ifx\csname urlstyle\endcsname\relax
  \providecommand{\doi}[1]{doi: #1}\else
  \providecommand{\doi}{doi: \begingroup \urlstyle{rm}\Url}\fi

\bibitem[Alpher(2002)]{Alpher02}
A.~Alpher.
\newblock Frobnication.
\newblock \emph{Journal of Foo}, 12\penalty0 (1):\penalty0 234--778, 2002.

\bibitem[Alpher and Fotheringham-Smythe(2003)]{Alpher03}
A.~Alpher and J.~P.~N. Fotheringham-Smythe.
\newblock Frobnication revisited.
\newblock \emph{Journal of Foo}, 13\penalty0 (1):\penalty0 234--778, 2003.

\bibitem[Alpher et~al.(2004)Alpher, Fotheringham-Smythe, and Gamow]{Alpher04}
A.~Alpher, J.~P.~N. Fotheringham-Smythe, and G.~Gamow.
\newblock Can a machine frobnicate?
\newblock \emph{Journal of Foo}, 14\penalty0 (1):\penalty0 234--778, 2004.

\bibitem[Authors(2006{\natexlab{a}})]{Authors06}
Authors.
\newblock The frobnicatable foo filter, 2006{\natexlab{a}}.
\newblock ECCV06 submission ID 324. Supplied as additional material {\tt
  eccv06.pdf}.

\bibitem[Authors(2006{\natexlab{b}})]{Authors06b}
Authors.
\newblock Frobnication tutorial, 2006{\natexlab{b}}.
\newblock Supplied as additional material {\tt tr.pdf}.

\bibitem[Mermin(1989)]{Mermin89}
N.~David Mermin.
\newblock What's wrong with these equations?
\newblock \emph{Physics Today}, October 1989.
\newblock \small\url{http://www.cvpr.org/doc/mermin.pdf}.

\end{thebibliography}
\end{document}